\documentclass[utf8]{FrontiersinHarvard} 

\usepackage{url,hyperref,lineno,microtype}
\usepackage{booktabs}
\usepackage[onehalfspacing]{setspace}

\def\keyFont{\fontsize{8}{11}\helveticabold }
\def\firstAuthorLast{Alexandre Bittar {et~al.}} 
\def\Authors{Alexandre Bittar\,$^{1,2,*}$, Philip N. Garner\,$^{2}$}
\def\Address{
  $^{1}$Idiap Research Institute, Martigny, Switzerland \\
  $^{2}$\'Ecole Polytechnique Fédérale de Lausanne, Lausanne, Switzerland
  }
\def\corrAuthor{Alexandre Bittar}
\def\corrEmail{abittar@idiap.ch}

\begin{document}
\onecolumn
\firstpage{1}

\title[Neural oscillations during speech perception via spiking neural networks]{Exploring neural oscillations during speech perception via surrogate gradient spiking neural networks} 

\author[\firstAuthorLast ]{\Authors} 
\address{} 
\correspondance{} 

\extraAuth{}

\maketitle

\begin{abstract}
Understanding cognitive processes in the brain demands sophisticated models capable of replicating neural dynamics at large scales. We present a physiologically inspired speech recognition architecture, compatible and scalable with deep learning frameworks, and demonstrate that end-to-end gradient descent training leads to the emergence of neural oscillations in the central spiking neural network. Significant cross-frequency couplings, indicative of these oscillations, are measured within and across network layers during speech processing, whereas no such interactions are observed when handling background noise inputs. Furthermore, our findings highlight the crucial inhibitory role of feedback mechanisms, such as spike frequency adaptation and recurrent connections, in regulating and synchronising neural activity to improve recognition performance. Overall, on top of developing our understanding of synchronisation phenomena notably observed in the human auditory pathway, our architecture exhibits dynamic and efficient information processing, with relevance to neuromorphic technology.

\tiny
 \keyFont{ \section{Keywords:} neural oscillations, spiking neural networks, speech recognition, brain-inspired computing, deep learning, surrogate gradient, spike-frequency adaptation, neuromorphic computing} 
\end{abstract}

\section{Introduction}

In the field of speech processing technologies, the effectiveness of training deep artificial neural networks (ANNs) with gradient descent has led to the emergence of many successful encoder-decoder architectures for automatic speech recognition (ASR), typically trained in an end-to-end fashion over vast amounts of data \citep{Gulati2020, Baevski2020, Li2021, Radford2023}. Despite recent efforts \citep{Brodbeck2024, Millet2022, Millet2021, Magnuson2020} towards understanding how these ANN representations can compare with speech processing in the human brain, the cohesive integration of the fields of deep learning and neuroscience remains a challenge. Nonetheless, spiking neural networks (SNNs), a type of artificial neural network inspired by the biological neural networks in the brain, present an interesting convergence point of the two disciplines. Although slightly behind in terms of performance compared to ANNs, SNNs have recently achieved concrete progress \citep{Hammouamri2023, Sun2023, Bittar2022a, Yin2021} on speech command recognition tasks using the surrogate gradient method \citep{Neftci2019} which allows them to be trained via gradient descent. Further work has also shown that they can be used to define a spiking encoder inside a hybrid ANN-SNN end-to-end trainable architecture on the more challenging task of large vocabulary continuous speech recognition \citep{Bittar2022c}. Their successful inclusion into contemporary deep learning ASR frameworks offers a promising path to bridge the existing gap between deep learning and neuroscience in the context of speech processing. This integration not only equips deep learning tools with the capacity to engage in speech neuroscience but also offers a scalable approach to simulate spiking neural dynamics, which supports the exploration and testing of hypotheses concerning the neural mechanisms and cognitive processes related to speech. This investigation of complex brain functions via physiologically inspired networks aligns with the work of \citet{Pulvermuller2021, Henningsen2022, Pulvermuller2023}, who applied biological constraints to large-scale simulations of language learning in SNNs. We complement their approach by training on more realistic speech data, albeit at the cost of some simplifications.

In neuroscience, various neuroimaging techniques such as electroencephalography (EEG) can detect rhythmic and synchronised postsynaptic potentials that arise from activated neuronal assemblies. These give rise to observable neural oscillations, commonly categorised into distinct frequency bands: delta (0.5-4 Hz), theta (4-8 Hz), alpha (8-13 Hz), beta (13-30 Hz), low-gamma (30-80 Hz), and high-gamma (80-150 Hz) \citep{Buzsaki2006}. It is worth noting that while these frequency bands provide a useful framework, their boundaries are not rigidly defined and can vary across studies. Nevertheless, neural oscillations play a crucial role in coordinating brain activity and are implicated in cognitive processes such as attention \citep{Fries2001, Jensen2007, Womelsdorf2007, Vinck2013}, memory \citep{Kucewicz2017}, sensory perception \citep{Bacsar2000, Senkowski2007} and motor function \citep{Mackay1997, Ramos2015}. Of particular interest is the phenomenon of cross-frequency coupling (CFC) which reflects the interaction between oscillations occurring in different frequency bands \citep{Jensen2007, Jirsa2013}. As reviewed in \citet{Abubaker2021}, many studies have demonstrated a relationship between CFC and working memory performance \citep{Tort2009, Axmacher2010}. In particular phase-amplitude coupling (PAC) between theta and gamma rhythms appears to support memory integration \citep{Buzsaki2013, Backus2016, Hummos2017}, preservation of sequential order \citep{Reddy2021, Colgin2013, Itskov2008} and information retrieval \citep{Mizuseki2009}. In contrast, alpha-gamma coupling commonly manifests itself as a sensory suppression mechanism during selective attention \citep{Foxe2011, Banerjee2011}, inhibiting task-irrelevant brain regions \citep{Jensen2010} and ensuring controlled access to stored knowledge \citep{Klimesch2012}. Finally, beta oscillations are commonly associated with cognitive control and top-down processing \citep{Engel2001}.

In the context of speech perception, numerous investigations have revealed a similar oscillatory hierarchy, where the temporal organisation of high-frequency signal amplitudes in the gamma range is orchestrated by low-frequency neural phase dynamics, specifically in the delta and theta ranges \citep{Canolty2006, Ghitza2011, Giraud2012, Hyafil2015, Attaheri2022}. These three temporal scales -- delta, theta and gamma -- naturally manifest in speech and represent specific perceptual units. In particular, delta-range modulation (1-2 Hz) corresponds to perceptual groupings formed by lexical and phrasal units, encapsulating features such as the intonation contour of an utterance. Modulation within the theta-range aligns with the syllabic rate (4 Hz) around which the acoustic envelope consistently oscillates. Finally, (sub)phonemic attributes, including formant transitions that define the fine structure of speech signals, correlate with higher modulation frequencies (30-50 Hz) within the low-gamma range. The close correspondence between the perception of (sub)phonemic, syllabic and phrasal attributes on one hand, and the manifestation of gamma, theta and delta neural oscillations on the other, was notably emphasised in \citet{Giraud2012}. These different levels of temporal granularity inherent to speech signals therefore appear to be processed in a hierarchical fashion, with the intonation and syllabic contour encoded by earlier neurons guiding the encoding of phonemic features by later neurons. Some insights about how phoneme features end up being encoded in the temporal gyrus were given in \citet{Mesgarani2014}. Drawing from recent research \citep{Bonhage2017} on the neural oscillatory patterns associated with the sentence superiority effect, it is suggested that such low-frequency modulation may facilitate automatic linguistic chunking by grouping higher-order features into packets over time, thereby contributing to enhanced sentence retention. The engagement of working memory in manipulating phonological information enables the sequential retention and processing of speech sounds for coherent word and sentence representations. Additionally, alpha modulation has also been shown to play a role in improving auditory selective attention \citep{Strauss2014a, Strauss2014b, Wostmann2017}, reflecting the listener's sensitivity to acoustic features and their ability to comprehend speech \citep{Obleser2012}.

Computational models \citep{Hyafil2015, Hovsepyan2020} have shown that theta oscillations can indeed parse speech into syllables and provide a reliable reference time frame to improve gamma-based decoding of continuous speech. These approaches \citep{Hyafil2015, Hovsepyan2020} implement specific models for theta and gamma neurons along with a distinction between inhibitory and excitatory neurons. The resulting networks are then optimised to detect and classify syllables with very limited numbers of trainable parameters (10-20). In contrast, this work proposes to utilise significantly larger end-to-end trainable multi-layered architectures (400k-20M trainable parameters) where all neuron parameters and synaptic connections are optimised to predict sequences of phoneme/subword probabilities, that can subsequently be decoded into words. By avoiding constraints on theta or gamma activity, the approach allows us to explore which forms of CFC naturally arise when solely optimising the decoding performance. Even though the learning mechanism is not biologically plausible, we expect that a model with sufficiently realistic neuronal dynamics and satisfying ASR performance should reveal similarities with the human brain. We divide our analysis in two parts, 
\begin{enumerate}
    \item \textbf{Architecture}: As a preliminary analysis, we conduct hyperparameter tuning to optimise the model's architectural parameters. On top of assessing the network's capabilities and scalability, we notably evaluate how the incorporation of spike-frequency adaptation (SFA) and recurrent connections impact the speech recognition performance. 
    \item \textbf{Oscillations}: We then explore the central aspect of our analysis concerning the emergence of neural oscillations within our model. Each SNN layer is treated as a distinct neuron population, from which spike trains are aggregated into a population signal similar to EEG data. Through intra- and inter-layer CFC analysis, we investigate the presence of significant delta-gamma, theta-gamma, alpha-gamma and beta-gamma PAC. We also investigate how incorporating Dale's law, SFA and recurrent connections affect the synchronisation of neural activity. 
\end{enumerate}

\newpage
\section{Materials and methods}

\subsection{Spiking neuron model}
\label{sec:neuron}

\begin{figure}[ht]
    \centering
    \includegraphics[width=0.7\columnwidth]{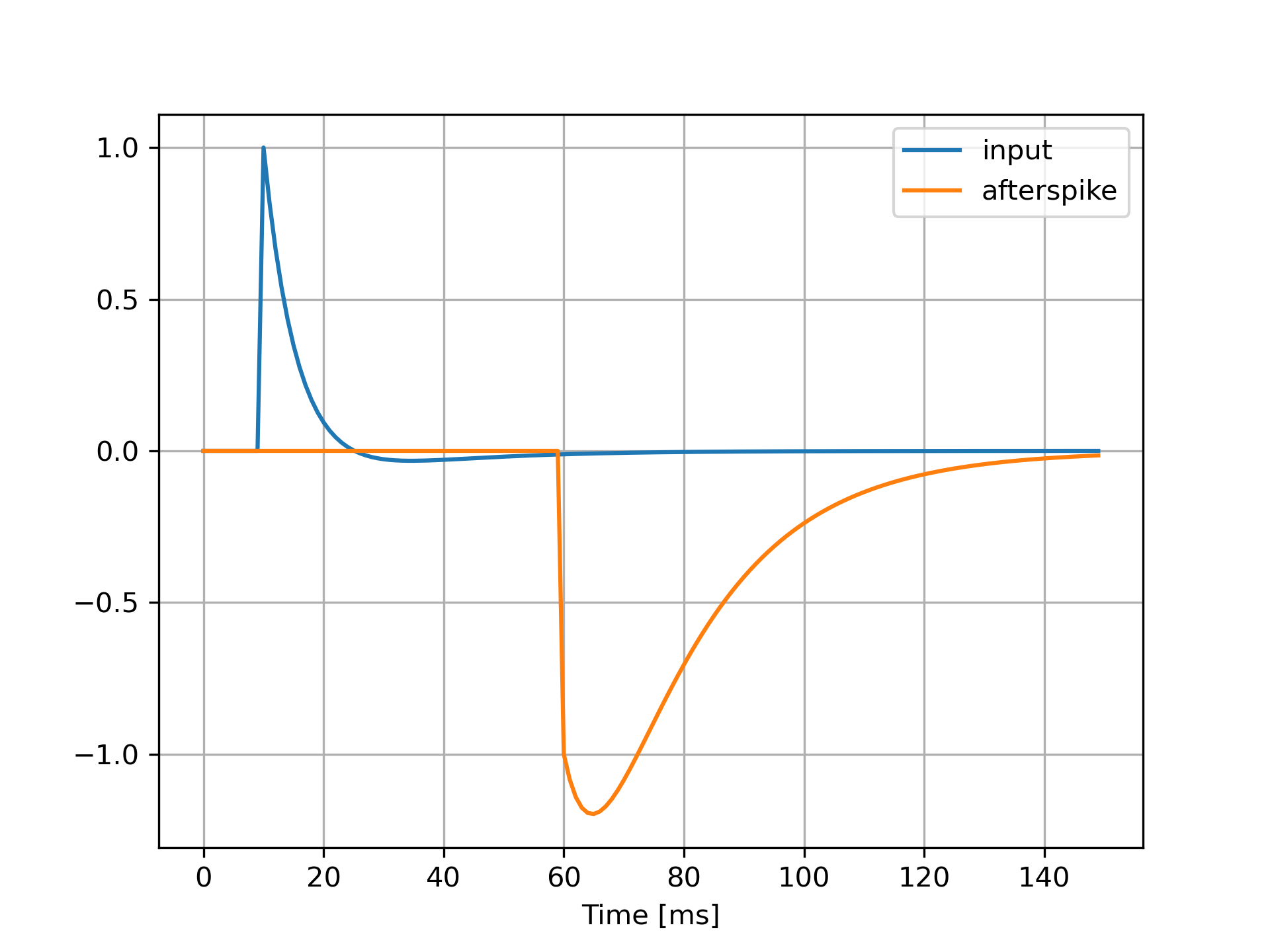}
    \caption{Kernel functions of AdLIF neuron model. Membrane potential response to an input pulse at $t=10$ ms (in blue) and to an emitted spike at $t=60$ ms (in orange). The neuron parameters are $\tau_u=5$ ms, $\tau_w=30$ ms, $a=0.5$ and $b=1.5$.}
    \label{fig:kernels} 
\end{figure}

Physiologically grounded neuron models such as the well known Hodgkin and Huxley model \citep{Hodgkin1952} can be reduced to just two variables \citep{Fitzhugh1961, MorrisLecar1981}. More contemporary models, such as the Izhikevich \citep{Izhikevich2003} and adaptive exponential integrate-and-fire \citep{Brette2005} models, have similarly demonstrated the capacity to accurately replicate voltage traces observed in biological neurons using just membrane potential and adaptation current as essential variables \citep{Badel2008}. With the objective of incorporating realistic neuronal dynamics into large-scale neural network simulations with gradient descent training, the linear AdLIF neuron model stands out as an adequate compromise between physiological plausibility and computational efficiency. It can be described in continuous time by the following differential equations \citep{Gerstner2002},
\begin{align}
    \label{adlif1}
    \tau_u\,\dot{u}(t)&=-\big(u(t)-u_{\text{rest}}\big) -R\, w(t) + R\,I(t) -\tau_u\,(\vartheta-u_r)\,\sum_f\delta(t-t^f) \\[+3pt]
    \label{adlif2}
    \tau_w\,\dot{w}(t)&=-w(t)+a\big(u(t)-u_{\text{rest}}\big) + \tau_w\,b \,\sum_f\delta(t-t^f)\, .
\end{align}
The neuron's internal state is characterised by the membrane potential $u(t)$ which linearly integrates stimuli $I(t)$ and gradually decays back to a resting value $u_{\text{rest}}$ with time constant $\tau_u\in[3, 25]$ ms. A spike is emitted when the threshold value $\vartheta$ is attained, $u(t)\geq\vartheta$, denoting the firing time $t=t^f$, after which the potential is decreased by a fixed amount $\vartheta-u_r$. In the following, we will use $u_r=u_{\text{rest}}$ for simplicity. The second variable $w(t)$ is coupled to the potential with strength $a$ and decay constant $\tau_w\in[30, 350]$ ms, characterising sub-threshold adaptation. Additionally, $w(t)$ experiences an increase of $b$ after a spike is emitted, which defines spike-triggered adaptation. The differential equations can be simplified as,
\begin{align}
    \label{adlif3}
    \tau_u\,\dot{u}(t)&=-u(t) -w(t) + I(t) -\tau_u\,\sum_f\delta(t-t^f) \\[+3pt]
    \label{adlif4}
    \tau_w\,\dot{w}(t)&=-w(t)+a\,u(t) + \tau_w\,b \,\sum_f\delta(t-t^f)\, .
\end{align}
by making all time-dependent quantities dimensionless with changes of variables,
\begin{equation*}
    u\rightarrow \frac{u-u_{\text{rest}}}{\vartheta-u_{\text{rest}}}\, , \quad w \rightarrow \frac{Rw}{\vartheta-u_{\text{rest}}}\quad \text{and} \quad I \rightarrow \frac{RI}{\vartheta-u_{\text{rest}}}    \, ,
\end{equation*}
and redefining neuron parameters as,
\begin{equation*}
    a\rightarrow Ra \, , \quad b\rightarrow \frac{R\,b}{\vartheta-u_{\text{rest}}} \, , \quad\vartheta\rightarrow\frac{\vartheta-u_{\text{rest}}}{\vartheta-u_{\text{rest}}}=1 \quad\text{and}\quad u_{\text{rest}} \rightarrow\frac{u_{\text{rest}}-u_{\text{rest}}}{\vartheta-u_{\text{rest}}}=0 \, .
\end{equation*}
This procedure gets rid of unnecessary parameters such as the resistance $R$, as well as resting, reset and threshold values, so that a neuron ends up being fully characterised by four parameters: $\tau_u$, $\tau_w$, $a$ and $b$. As derived in the Supplementary Appendix, the differential equations can be solved in discrete time with step size $\Delta t$ using a forward-Euler first-order exponential integrator method. After initialising $u_0=w_0=s_0=0$, and defining $\alpha:=\exp\frac{-\Delta t}{\tau_u}$ and $\beta:=\exp\frac{-\Delta t}{\tau_w}$, the neuronal dynamics can be solved by looping over time steps $t=1,2,\hdots,T$ as,
\begin{align}
    \label{eq_ut}
    u_t&=\alpha\big(u_{t-1}-s_{t-1}\big)+(1-\alpha)\big(I_t-w_{t-1}\big) \\
    \label{eq_wt}
    w_t&=\beta\big(w_{t-1}+b\,s_{t-1}\big)+(1-\beta)\,a\, u_{t-1} \\
    \label{eq_st}
    s_t&=\big(u_t\geq 1\big)\, .
\end{align}
Stability conditions for the value of the coupling strength $a$ are derived in the Supplementary Appendix. Additionally, we constrain the values of the neuron parameters to biologically plausible ranges \citep{Gerstner2002, Augustin2013},
\begin{equation}
    \tau_u\in[3,25]\,\text{ms}, \quad \tau_w\in[30,350]\,\text{ms}, \quad a\in[-0.5,5], \quad b\in[0,2] \, .
\end{equation}
This AdLIF neuron model is equivalent to a Spike Response Model (SRM) with continuous time kernel functions illustrated in Fig. \ref{fig:kernels}. The four neuron parameters $\tau_u$, $\tau_w$, $a$ and $b$ all characterise the shape of these two curves representing the membrane potential response to an input spike and to an emitted spike respectively. A derivation of the kernel-based SRM formulation is presented in the Supplementary Appendix.

\subsubsection{Spiking layers}

The spiking dynamics described in Eqs. \eqref{eq_ut} to \eqref{eq_st} can be computed in parallel for a layer of neurons. In a multi-layered network, the $l$-th layer receives as inputs a linear combination of the previous layer's outputs $s^{l-1}\in\{0,1\}^{B\times T\times N^{l-1}}$ where $B$, $T$ and $N^{l-1}$ represent the batch size, number of time steps and number of neurons respectively. Feedback from the $l$-th layer can also be implemented as a linear combination of its own outputs at the previous time step $s_{t-1}^{l}\in\{0,1\}^{N^{l}}$, so that the overall neuron stimulus $I^l_t$ for neurons in the $l$-th layer at time step $t$ is computed as,
\begin{equation}
    I_t^l=W^l\,s_t^{l-1}+V^l\,s_{t-1}^{l} \, .
\end{equation}
Here the feedforward $W^l\in\mathbb{R}^{N^{l-1}\times N^l}$ and feedback connections $V^l\in\mathbb{R}^{N^l\times N^l}$ are trainable parameters. Diagonal elements of $V^l$ are set to zero as afterspike self inhibition is already taken into account in Eq. \eqref{eq_ut}. While this choice excludes autapses, which are rare but do exist in the brain, it simplifies the model for our study. Additionally, a binary mask can be applied to matrices $W^l$ and $V^l$ to limit the number of nonzero connections. Similarly, a portion of neurons in a layer can be reduced to leaky integrate-and-fire (LIF) dynamics without any SFA by applying another binary mask to the neuron adaptation parameters $a$ and $b$. Indeed, if $a=b=0$, the adaptation current vanishes $w_t=0$ $\forall t\in\{1,2,\hdots,T\}$ and has no more impact on the membrane potential. 

\subsubsection{Surrogate gradient method}

The main challenge in applying stochastic gradient descent to the derived neuronal dynamics stems from the threshold operation described in Eq. \eqref{eq_st}, which has a derivative of zero everywhere except at the threshold point, where it is undefined. To address this, a surrogate derivative can be manually specified using PyTorch \citep{Paszke2017}, which enables the application of the Back-Propagation Through Time algorithm for training the resulting SNN  in a manner similar to recurrent neural network (RNN) training. In this paper, we adopt the boxcar function as our surrogate function. This choice has been proven effective in various contexts \citep{Kaiser2020, Bittar2022a, Bittar2022c} and requires minimal computational resources as expressed by the derivative definition,
\begin{equation}
    \frac{\partial s_t}{\partial u_t}=
    \begin{cases}
    0.5 \quad&\text{if}\quad |u_t-1|\leq 0.5 \\
    0 \quad&\text{otherwise}\, .
    \end{cases}
\end{equation}

\subsubsection{Spike frequency regularisation}

The firing rate $f^l_{b,n}$ of neuron $n$ in layer $l$ when processing utterance $b$ can be calculated in Hz as,
\begin{equation}
    f_{b,n}^l=\frac{1}{T_b}\sum_{t=1}^T s^l_{b,t,n} \, ,
\end{equation}
where $T_b$ is the utterance duration in seconds. We regularise the firing rates of all spiking neurons between $f_{\text{min}}=0.5$ Hz and $f_{\text{max}}=f_{\text{Nyquist}}$ using the following regularisation loss, 
\begin{equation}
\label{eq_reg}
    \mathcal{L}_{\text{reg}}=\frac{1}{B\,L}\sum_{b=1}^B\sum_{l=1}^L\frac{1}{N^l}\sum_{n=1}^{N^l}\text{ReLU}\Big(f_{\text{min}}-f^l_{b,n}\Big)+\text{ReLU}\Big(f^l_{b,n}-f_{\text{max}}\Big) \, ,
\end{equation}
to discourage neurons from remaining silent or from firing above the Nyquist frequency.

\subsection{Overview of the auditory pathway in the brain}

Sound waves are initially received by the outer ear and then transmitted as vibrations to the cochlea in the inner ear, where the basilar membrane allows for a representation of different frequencies along its length \citep{Gundersen1978}. Distinct sound frequencies induce localised membrane vibrations that activate adjacent inner hair cells. These specialised sensory cells, covering the entire basilar membrane, release neurotransmitters when activated, stimulating neighbouring auditory nerve fibers and initiating the production of action potentials. Tonotopy is maintained through the conversion of mechanical motion into electric signals as each inner hair cell, tuned to a specific frequency, only affects nearby auditory nerve fibers \citep{Saenz2014}. The resulting spike trains then propagate through a multi-layered neural network, ultimately reaching cortical regions associated with higher-order cognitive functions such as speech recognition. Overall, the auditory system is organised hierarchically, with each level contributing to the progressively more sophisticated processing of auditory information.

\subsection{Simulated speech recognition pipeline}

\begin{figure}[ht]
    \centering
    \includegraphics[width=1.0\columnwidth]{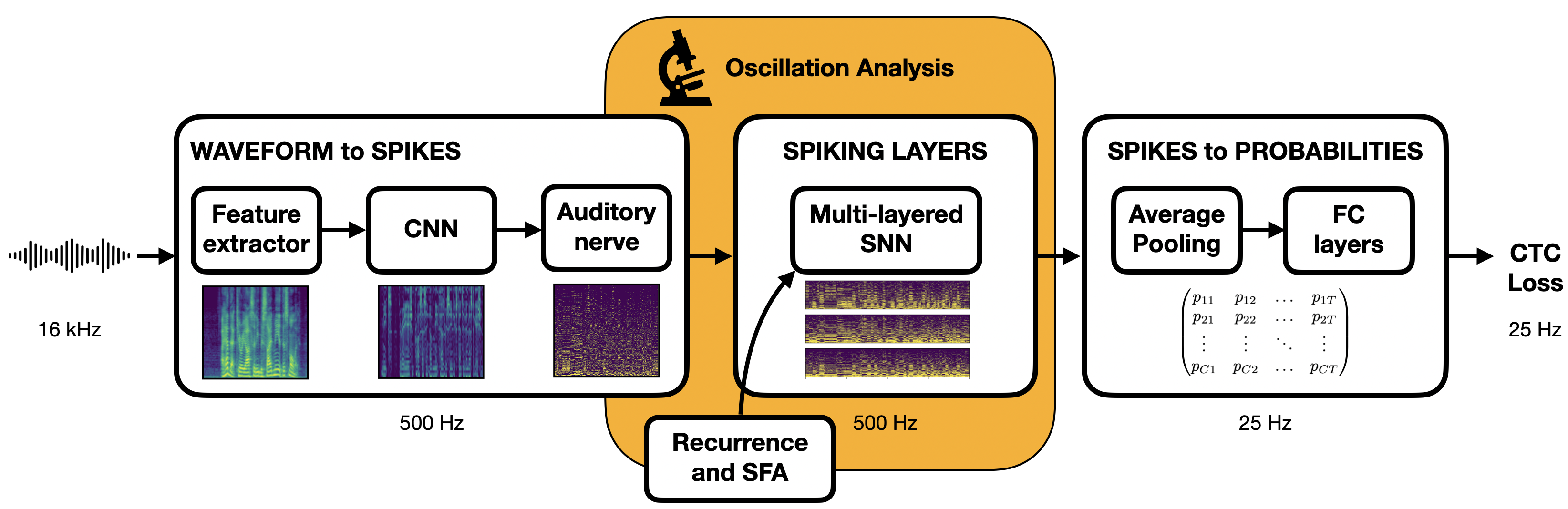}
    \caption{End-to-end trainable speech recognition pipeline. Input waveform is converted to a spike train representation to be processed by the central SNN before being transformed into output phoneme probabilities sent to a loss function for training.}
    \label{fig:pipeline}
\end{figure}

Our objective is to design a speech recognition architecture that, while sufficiently plausible for meaningful comparisons with neuroscience observations, remains simple and efficient to ensure compatibility with modern deep learning techniques and achieve good ASR performance. We implement the overall waveform-to-phoneme pipeline illustrated in Fig. \ref{fig:pipeline} inside the Speechbrain \citep{speechbrain} framework. We provide a description of each of its components here below.

\subsubsection{Feature extractor}
Mel filterbank features are extracted from the raw waveform using 80 filters and a 25 ms window with a shift of 2 ms. This procedure down samples the 16 kHz input speech signal to a 500 Hz spectrogram representation with 80 frequency bins.

\subsubsection{Auditory CNN}
A single-layered two-dimensional convolution module is applied to the 80 extracted Mel features using 16 channels, a kernel size of (7, 7), a padding of (7, 0) and a stride of 1, producing $16\cdot\big((80-7)+1\big)=1184$ output signals with unchanged number of time steps. Layer normalisation, drop out on the channel dimension and a Leaky-ReLU activation are then applied. Each produced signal characterises the evolution over time of the spectral energy across a frequency band of 7 consecutive Mel bins.

\subsubsection{Auditory nerve fibers}
Each 500 Hz output signal from the auditory CNN constitutes the stimulus of a single auditory nerve fiber, which converts the real-valued signal into a spike train. These nerve fibers are modelled as a layer of LIF neurons without recurrent connections and using a single trainable parameter per neuron, $\tau_u\in[3,25]$ ms, representing the time constant of the membrane potential decay.

\subsubsection{Multi-layered SNN}
The resulting spike trains are sent to a fully connected multi-layered SNN architecture with 512 neurons in each layer. The proportion of neurons with nonzero adaptation parameters is controlled in each layer so that only a fraction of the neurons are AdLIF and the rest are LIF. Similarly the proportion of nonzero feedforward and recurrent connections is controlled in each layer by applying fixed random binary masks to the weight matrices. Compared to a LIF neuron, an AdLIF neuron has three additional trainable parameters, $\tau_w\in[30, 350]$ ms, $a\in [-0.5,5]$ and $b\in [0,2]$, related to the adaptation variable coupled to the membrane potential.

\subsubsection{Spikes to probabilities}
The spike trains of the last layer are sent to a an average pooling module which down samples their time dimension to 25 Hz. These are then projected to 512 phoneme features using two fully connected (FC) layers with Leaky-ReLU activation. A third FC layer with Log-Softmax activation finally projects them to 40 log-probabilities representing 39 phoneme classes and a blank token as required by connectionist temporal classification (CTC).

\subsubsection{Training and inference}
The log-probabilities are sent to a CTC loss \citep{Graves2006} so that the parameters of the complete architecture can be updated through back propagation. Additionally, regularisation of the firing rate as defined in Eq. \eqref{eq_reg} is used to prevent neurons from being silent or firing above the Nyquist frequency. At inference, CTC decoding is used to output the most likely phoneme sequence from the predicted log-probabilities, and the phoneme error rate (PER) is computed to evaluate the model's performance.

\subsection{Physiological plausibility and limitations}

\subsubsection{Cochlea and inner hair cells}
While some of the complex biological processes involved in converting mechanical vibrations to electric neuron stimuli can be abstracted, we assume that the key feature to retain is the tonotopic encoding of sound information. A commonly used metric in neuroscience is the ratio of characteristic frequency to bandwidth, which defines how sharply tuned a neuron is around the frequency it is most responsive to. As detailed in Supplementary Table S6, from measured Q10 values in normal hearing humans reported in \citet{Devi2022}, we evaluate that a single auditory nerve fiber should receive inputs from 5-7 adjacent frequency bins when using 80 Mel filterbank features. The adoption of a Mel filterbank frontend can be justified by its widespread utilisation within deep learning ASR frameworks. Although we do not attempt to directly model cochlear and hair cell processing, we can provide a rough analogue in the form of Mel features passing through a trainable convolution module that yields plausible ranges of frequency sensitivity for our auditory nerve fibers.

\subsubsection{Simulation time step}
Modern ASR systems \citep{Gulati2020, Radford2023} typically use a frame period of $\Delta t=10$ ms during feature extraction, which is then often sub-sampled to 40 ms using a CNN before entering the encoder-decoder architecture. In the brain, typical minimal inter-spike distances imposed by a neuron's absolute refractory period can vary from 0 to 5 ms \citep{Gerstner2002}. We therefore assume that using a time step greater than 5 ms could result in dynamics that are less representative of biological phenomena. Although using a time step $\Delta t<1$ ms may yield biologically more realistic simulations, we opt for time steps ranging from 1 to 5 ms to ensure computational efficiency. After the SNN, the spike trains of the last layer are down-sampled to 25 Hz via average pooling on the time dimension. This prevents an excessive number of time steps from entering the CTC loss, which could potentially hinder its decoding efficacy. We use $\Delta t=5$ ms for most of the hyperparameter tuning to reduce training time, but favour $\Delta t=2$ ms for the oscillation analysis so that the full gamma range of interest (30-150 Hz) remains below the Nyquist frequency at 250 Hz.

\subsubsection{Neuron model}
The LIF neuron model is an effective choice for modelling auditory nerve fibers as it accurately represents their primary function of encoding sensory inputs into spike trains. We avoid using SFA and recurrent connections, as they are not prevalent characteristics of nerve fibers. On the other hand, for the multi-layered SNN, the linear AdLIF neuron model with layer-wise recurrent connections stands out as a good compromise between accurately reproducing biological firing patterns and remaining computationally efficient \citep{Bittar2022a, Deckers2024}. Although less popular than the moving threshold formulation by \citet{Bellec2018}, recently reviewed in \citet{Ganguly2024}, our implementation of SFA using the AdLIF model combines spike-triggered adaptation with subthreshold coupling. Previous work \citep{Bittar2022a} demonstrated that the AdLIF outperforms moving threshold implementations \citep{Yin2021, Salaj2021, Shaban2021, Yin2020} in speech command recognition tasks. Nevertheless, we will still implement and train an additional model with moving threshold SFA to ensure that our conclusions hold consistently across different SFA models.

\subsubsection{Organisation in layers}
Similarly to ANNs, our simulation incorporates a layered organisation, which facilitates the progressive extraction and representation of features from low-order to higher-order, without the need of concretely defining and distinguishing neuron populations. This fundamental architectural principle aligns with the general hierarchical processing observed in biological brains. However, it oversimplifies the complexities of auditory processing, which extends beyond a straightforward sequential framework. While there is some sort of sequential processing in sub-cortical structures, the levels of processed features are more intricate than a simple hierarchy. This simplification is made to ensure compatibility with deep learning frameworks. 

\subsubsection{Layer-wise recurrence}
While biological efferent pathways in the brain involve complex and widespread connections that span across layers and regions, modelling such intricate connectivity can introduce computational challenges and complexity, potentially hindering training and scalability. By restricting feedback connections to layer-wise recurrence, we simplify the network architecture and enhance compatibility with deep learning frameworks.

\subsubsection{Excitatory and inhibitory}
In the neuroscience field, neurons are commonly categorised into two types: excitatory neurons, which stimulate action potentials in postsynaptic neurons, and inhibitory neurons, which reduce the likelihood of spike production in postsynaptic neurons. This principle is referred to as Dale's law.

In ANNs, weight matrices are commonly initialised with zero mean and a symmetric distribution, so that the initial number of excitatory and inhibitory connections is balanced. During training, synaptic connections are updated across all layers without enforcing a distinction between excitatory and inhibitory neurons. Dale's law can nevertheless be imposed \citep{Li2024, Cornford2021} even if it typically results in slightly reduced performance.

In our baseline model, Dale's law is not applied, so that similarly to standard ANNs, weight matrices are trained without constraining values to be positive or negative. Additionally, we train separate SNNs with Dale's law to evaluate its impact on neural oscillations. In this setup, half the neurons are excitatory and half are inhibitory, while the auditory nerve fibers are all excitatory.

\subsubsection{Delays}

In biological neural networks, the propagation time of spikes between neurons introduces delays, primarily due to axonal transmission. To incorporate and assess the impact of these delays on neural oscillations, we additionally train separate SNNs using dilated convolutions in the temporal dimension instead of fully connected feedforward matrices. This approach, introduced in \citet{Hammouamri2023}, allows us to introduce controlled delays directly into the network architecture. Based on their configuration for speech command recognition tasks, we use a maximum delay value of 300 ms.

\subsubsection{Learning rule}
Stochastic gradient descent, though biologically implausible due to its global and offline learning framework, allows us to leverage parallelisable and fast computations to optimise larger-scale neural networks. While this approach facilitates effective training and scaling, it diverges from biologically inspired synaptic plasticity mechanisms, such as those mediated by AMPA and NMDA receptors.

\subsubsection{Decoding into phoneme sequences}
Although lower PERs could be achieved with a more sophisticated decoder, our primary focus is on analysing the spiking layers within the encoder. For simplicity, we therefore opt for straightforward CTC decoding, which more directly reflects the encoder's capabilities.

\subsubsection{Hybrid ANN-SNN balance}

The CNN module in the ASR frontend as well as the ANN module (average pooling and FC layers) converting spikes to probabilities are intentionally kept simple to give most of the processing and representational power to the central SNN on which focuses our neural oscillations analysis.

\subsection{Speech processing tasks}
\label{sec:tasks}

The following datasets are used in our study.
\begin{itemize}
    \item The TIMIT dataset \citep{Garofolo1993} provides a comprehensive and widely utilised collection of phonetically balanced American English speech recordings from 630 speakers with detailed phonetic transcriptions and word alignments. It represents a standardised benchmark for evaluating ASR model performance. The training, validation and test sets contain 3696, 400 and 192 sentences respectively. Utterance durations vary between 0.9 to 7.8 seconds. Due to its compact size of approximately five hours of speech data, the TIMIT dataset is well-suited for investigating suitable model architectures and tuning hyperparameters. It is however considered small for ASR hence the use of Librispeech presented below.

    \item The Librispeech corpus \citep{Panayotov2015} contains about 1,000 hours of English speech audiobook data read by over 2,400 speakers with utterance durations between 0.8 and 35 seconds. Given its significantly larger size, we only train a few models selected on TIMIT to confirm that our analysis holds when scaling to more data.

    \item The Google Speech Commands dataset \citep{Warden2018} contains one-second audio recordings of 35 spoken commands such as "yes," "no," "stop," "go," "left," "right" "up". The training, validation and testing splits contain approximately 85k, 10k and 5k examples respectively. It is used to test whether similar CFCs arise when simply recognising single words instead of phoneme or subword sequences.
\end{itemize}

When evaluating an ASR model, the error rate signifies the proportion of incorrectly predicted words or phonemes. The count of successes in a binary outcome experiment, such as ASR testing, can be effectively modeled using a binomial distribution. In the context of trivial priors, the posterior distribution of the binomial distribution follows a beta distribution. By leveraging the equal-tailed 95$\%$ credible intervals derived from the posterior distribution, we establish error bars, yielding a range of $\pm0.8\%$ for the reported PERs on the TIMIT dataset, about $\pm0.2\%$ for the reported word error rates on LibriSpeech, and about $\pm0.4\%$ for the reported accuracy on Google Speech Commands.

\subsection{Analysis methods}
\label{sec:analysis}

\subsubsection{Hyperparameter tuning}
Before reporting results on the oscillation analysis, we investigate the optimal architecture by tuning some relevant hyperparameters. All experiments are run using a single NVIDIA GeForce RTX 3090 GPU. On top of assessing their respective impact on the error rate, we test if more physiologically plausible design choices correlate with better performance. Here is the list of the fixed parameters that we do not modify in our reported experiments:
\begin{itemize}[noitemsep]
    \item number of Mel bins: 80
    \item Mel window size: 25 ms
    \item auditory CNN kernel size (7, 7)
    \item auditory CNN stride: (1, 1)
    \item auditory CNN padding: (7, 0)
    \item average pooling size: 40 / $\Delta t$
    \item number of phoneme FC layers: 2 
    \item number of phoneme features: 512
    \item dropout: 0.15
    \item activation: LeakyReLU
\end{itemize}
where for CNN attributes of the form $(n_t, n_f)$, $n_t$ and $n_f$ correspond to time and feature dimensions respectively. The tunable parameters are the following,
\begin{itemize}[noitemsep]
    \item filter bank hop size controlling the SNN time step in ms: \{1, 2, 5\}
    \item number of auditory CNN channels (filters): \{8, 16, 32, 64, 128\}
    \item number of SNN layers: \{1, 2, 3, 4, 5, 6, 7\}
    \item neurons per SNN layer: \{64, 128, 256, 512, 768, 1024, 1536, 2048\}
    \item proportion of neurons with SFA: [0, 1]
    \item feedforward connectivity: [0, 1]
    \item recurrent connectivity: [0, 1]
\end{itemize}
While increasing the number of neurons per layer primarily impacts memory requirements, additional layers mostly extend training time.

\subsubsection{Population signal}
In the neuroscience field, EEG stands out as a widely employed and versatile method for studying brain activity. By placing electrodes on the scalp, this non-invasive technique measures the aggregate electrical activity resulting from the synchronised firing of neurons within a specific brain region. An EEG signal therefore reflects the summation of postsynaptic potentials from a large number of neurons operating in synchrony. The typical sampling rate for EEG data is commonly in the range of 250 to 1000 Hz which matches our desired simulation time steps. With our SNN, we do not have EEG signals but directly the individual spike trains of all neurons in the architecture. In order to perform similar population-level analyses, we sum the binary spike trains $s^l_b \in \{0,1\}^{T\times N^l}$ emitted by all neurons in a specific layer $l$ for a single utterance $b$ as follows,
\begin{equation}
    p^l_{b,t}=\sum_{n=1}^{N^l}s^l_{b,t,n} \, .
\end{equation}
Before performing the PAC analysis, the resulting population activity signal $p^l_{b,t}$ is then normalised over the time dimension with a mean of 0 and a standard deviation of 1, yielding the normalised population signal $\hat{p}^l_{b,t}$ defined as,
\begin{equation}
    \hat{p}^l_{b,t} = \frac{p^l_{b,t} - \mu^l_b}{\sigma^l_b} \, ,
\end{equation}
where $\mu^l_b$ is the mean and $\sigma^l_b$ is the standard deviation of $p^l_{b,t}$ over the time dimension.

\subsubsection{Phase-amplitude coupling}

Using finite impulse response band-pass filters, the obtained population signals are decomposed into different frequency ranges. We study CFC in the form of PAC both within a single population and across layers. This technique assesses whether a relationship exists between the phase of a low frequency signal and the envelope (amplitude) of a high frequency signal. 

As recommended in \citet{Hulsemann2019}, we implement both the modulation index \citep{Tort2008} and mean vector length \citep{Canolty2006} metrics to quantify the observed amount of PAC. For each measure type, the observed coupling value is compared to a distribution of 10,000 surrogates to assess the significance. Surrogate couplings are computed by disrupting the temporal order of the amplitude time series while preserving its overall characteristics. Specifically, the amplitude time series is permuted by cutting it at a random point and reversing the order of the two segments. This method, as discussed by \citet{Hulsemann2019}, maintains all inherent properties of the original data except for the temporal relationship between phase angle and amplitude magnitude, providing a conservative test of significance. A p-value is then obtained by fitting a Gaussian function on the distribution of surrogate coupling values and calculating the area under the curve for values greater than the observed coupling value. 

As pointed out in \citet{Jones2016}, it is important to note that observed oscillations can exhibit complexities such as non-sinusoidal features and brief transient events on single trials. Such nuances become aggregated when averaging signals, leading to the widely observed continuous rhythms. We therefore perform all analysis on single utterances. 

For intra-layer interactions, a single population signal is used to extract both the low-frequency oscillation phase and the high-frequency oscillation amplitude. In a three-layered architecture, these interactions include nerve-nerve, first layer-first layer, second layer-second layer, and third layer-third layer couplings. 

For inter-layer interactions, we consider couplings between the low-frequency oscillation phase in one layer and the high-frequency oscillation amplitude in all subsequent layers. These interactions include nerve-first layer, nerve-second layer, nerve-third layer, first layer-second layer, first layer-third layer, second layer-third layer couplings. 

For all aforementioned intra- and inter-layer combinations, we use delta (0.5-4 Hz), theta (4-8 Hz), alpha (8-13 Hz) and beta (13-30 Hz) ranges as low-frequency modulating bands, and low-gamma (30-80 Hz) and high-gamma (80-150 Hz) ranges as high-frequency modulated bands. For a given model, we iterate through the 64 longest utterances in the TIMIT test set. For each utterance, we consider the 10 aforementioned intra- and inter-layer relations, as well as the 8 possible combinations of low-frequency to high-frequency bands. We conduct PAC testing on each of the 5,120 resulting coupling scenarios, and only consider a coupling to be significant when both modulation index and mean vector length metrics yield p-values below 0.05.

\newpage
\section{Results}

\subsection{Architectural analysis}

In order to draw a comparison with the human auditory pathway, we have introduced the physiologically inspired ASR pipeline illustrated in Fig. \ref{fig:pipeline}. The proposed hybrid ANN-SNN architecture is trained in an end-to-end fashion on the TIMIT dataset \citep{Garofolo1993} to predict phoneme sequences from speech waveforms. In the architectural design, we aimed to minimise the complexity of ANN components and favour the central SNN which will be the focus of the oscillation analysis. Here as a preliminary step, we examine how relevant hyperparameters affect the PER. On top of assessing the scalability of our approach to larger networks, we identify the importance of the interplay between recurrence and SFA. 

\subsubsection{Network scalability}
As reported in Table \ref{table:timit_layers}, performance improves with added layers, peaking at 4-6 layers before declining, which suggests a significant contribution to the final representation from all layers within this range. Compared to conventional non-spiking RNN encoders used in ASR, our results support the scalability of surrogate gradient SNNs to relatively deep architectures. Additionally, augmenting the number of neurons until about 1,000 per layer consistently yields lower PERs, beyond which performance saturates.

\begin{table}[ht]
\caption{Hyperparameter tuning for the number of SNN layers and neurons per layer on the TIMIT dataset. The third column gives both the number of trainable parameters in the multi-layered SNN (left) and in the whole encoder (right). The PERs are reported after 50 training epochs using a 5 ms time step, 16 CNN channels, 50\% of AdLIF neurons, 100\% feedforward and 50\% recurrent connectivity. The performance of the architecture when removing the SNN is also reported (bottom). Bold values indicate the lowest achieved PERs.}
\label{table:timit_layers}
\vspace{0.5cm}
\centering
\begin{tabular}{ccccc}
\toprule
\textbf{\begin{tabular}[c]{@{}c@{}}Number of\\ layers\end{tabular}} &
\textbf{\begin{tabular}[c]{@{}c@{}}Number of neurons \\ per layer\end{tabular}} &
\textbf{\begin{tabular}[c]{@{}c@{}}Number of\\ parameters\end{tabular}} &
\textbf{\begin{tabular}[c]{@{}c@{}}Test PER \\ {[}\%{]}\end{tabular}} &
\textbf{\begin{tabular}[c]{@{}c@{}}Validation PER \\ {[}\%{]}\end{tabular}} \\ \midrule
1 & 512 & 740k / 1.3M & 23.3 & 21.8 \\
2 & 512 & 1.1M / 1.7M & 21.0 & 19.2 \\
3 & 512 & 1.5M / 2.1M & 20.5 & 18.2 \\
4 & 512 & 1.9M / 2.5M & 20.2 & \textbf{17.4} \\
5 & 512 & 2.3M / 2.9M & \textbf{20.0} & 17.6 \\
6 & 512 & 2.7M / 3.3M & \textbf{20.0} & 17.9 \\
7 & 512 & 3.1M / 3.7M & 20.5 & 18.0 \\[+0.5ex]
\hline & \\[-1.5ex]
3 & 64 & 91k / 394k & 30.9 & 29.6 \\
3 & 128 & 211k / 547k & 25.5 & 24.1 \\
3 & 256 & 537k / 938k & 22.5 & 20.9 \\
3 & 768 & 3.0M / 3.7M & 19.6 & 17.4 \\
3 & 1024 & 4.9M / 5.7M & 19.1 & \textbf{17.1} \\
3 & 1536 & 10.1M / 11.2M & \textbf{19.0} & 17.3 \\
3 & 2048 & 17.1M / 18.5M & 19.2 & 17.2 \\[+0.5ex]
\hline & \\[-1.5ex]
\multicolumn{2}{c}{no nerve, no SNN} & 0 / 873k & 34.2 & 32.0 \\
\bottomrule
\end{tabular}
\end{table}

\subsubsection{Recurrent connections and spike-frequency adaptation}

The impact of adding SFA in the neuron model as well as using recurrent connections are reported in Table \ref{table:timit_rnn_sfa}. Interestingly, we find that without SFA, optimal performance is achieved by limiting the recurrent connectivity to 80$\%$. When additionally using SFA, further limitation of the recurrent connectivity about 50$\%$ yields the lowest PER. This differs from conventional non-spiking RNNs, where employing FC recurrent matrices is favoured. These results indicate that while requiring fewer parameters, an architecture with sparser recurrent connectivity and more selective parameter usage can achieve better task performance. 

Overall, SFA and recurrent connections individually yield significant error rate reduction, although they respectively grow as $\mathcal{O}(N)$ and $\mathcal{O}(N^2)$ with the number of neurons $N$. In line with previous studies on speech command recognition tasks \citep{Perez2021, Bittar2022a}, our results emphasise the metabolic and computational efficiency gained by harnessing the heterogeneity of adaptive spiking neurons. Furthermore, effectively calibrating the interplay between unit-wise feedback from SFA and layer-wise feedback from recurrent connections appears crucial for achieving optimal performance.

\begin{table}[ht]
\caption{Ablation experiments for the recurrent connectivity and proportion of neurons with SFA on the TIMIT dataset. PERs are reported after 50 epochs using a 5 ms time step, 16 CNN channels, 3 layers, 512 neurons per layer and 100\% feedforward connectivity. Bold values indicate the lowest achieved PERs.}
\label{table:timit_rnn_sfa}
\vspace{0.5cm}
\centering
\resizebox{\textwidth}{!}{
\begin{tabular}{lccccc}
\toprule
\textbf{Model configuration} & \textbf{\begin{tabular}[c]{@{}c@{}}Recurrent\\ connectivity\end{tabular}} &
\textbf{\begin{tabular}[c]{@{}c@{}}Proportion of\\ AdLIF neurons\end{tabular}} & \textbf{\begin{tabular}[c]{@{}c@{}}Number of \\ parameters\end{tabular}} &
\textbf{\begin{tabular}[c]{@{}c@{}}Test PER \\ {[}\%{]}\end{tabular}} &
\textbf{\begin{tabular}[c]{@{}c@{}}Validation PER \\ {[}\%{]}\end{tabular}} \\ \midrule
No recurrence no SFA & 0 & 0 & 1.1M / 1.7M & 26.9 & 24.8 \\[+0.5ex]
\hline & \\[-1.5ex]
Recurrence only & 0.2 & 0 & 1.3M / 1.8M & 22.0 & 20.1 \\
 & 0.5 & 0 & 1.5M / 2.1M & 21.0 & 18.9 \\
& 0.8 & 0 & 1.8M / 2.3M & 20.8 & 18.7 \\
& 1 & 0 & 1.9M / 2.5M & 21.8 & 19.3 \\[+0.5ex]
\hline & \\[-1.5ex]
SFA only & 0 & 0.2 & 1.1M / 1.7M & 24.2 & 21.7 \\
 & 0 & 0.5 & 1.1M / 1.7M & 23.7 & 21.6 \\
& 0 & 0.8 & 1.1M / 1.7M & 23.3 & 21.0 \\
& 0 & 1 & 1.1M / 1.7M & 22.9 & 21.1 \\[+0.5ex]
\hline & \\[-1.5ex]
Recurrence and SFA & 0.2 & 0.2 & 1.3M / 1.8M & 20.9 & 19.3 \\
 & 0.5 & 0.5 & 1.5M / 2.1M & \textbf{20.5} & \textbf{18.2} \\
& 0.8 & 0.8 & 1.8M / 2.3M & 21.2 & 18.8 \\
& 1 & 1 & 1.9M / 2.5M & 23.3 & 21.5 \\
\bottomrule
\end{tabular}
}
\end{table}

\subsubsection{Enforcing Dale's law}

To align with common ANN practice, the previous results were obtained without restricting neurons to be either strictly excitatory or strictly inhibitory. We now train more physiologically inspired models that satisfy Dale's law, with results presented in Table \ref{table:dale}. Although ASR performance decreased (1-4\% absolute PER increase), this may simply be due to suboptimal weight initialisation, which is known to affect performance \citep{Li2024}. This could likely be mitigated in future work by using the approach of \citet{Rossbroich2022}, who derived fluctuation-driven initialisation schemes compatible with Dale's law.

\begin{table}[ht]
\caption{Ablation experiments for the recurrent connectivity and proportion of neurons with SFA on the TIMIT dataset when additionally using Dale's law. PERs are reported after 50 epochs using a 2 ms time step, 16 CNN channels, 3 layers, 512 neurons per layer, 100\% feedforward connectivity and an excitatory-inhibitory ratio of 1. Bold values indicate the lowest achieved PERs.}
\label{table:dale}
\vspace{0.5cm}
\centering
\resizebox{\textwidth}{!}{
\begin{tabular}{lccccc}
\toprule
\textbf{Model configuration} & \textbf{\begin{tabular}[c]{@{}c@{}}Recurrent\\ connectivity\end{tabular}} &
\textbf{\begin{tabular}[c]{@{}c@{}}Proportion of\\ AdLIF neurons\end{tabular}} & \textbf{\begin{tabular}[c]{@{}c@{}}Number of \\ parameters\end{tabular}} &
\textbf{\begin{tabular}[c]{@{}c@{}}Test PER \\ {[}\%{]}\end{tabular}} &
\textbf{\begin{tabular}[c]{@{}c@{}}Validation PER \\ {[}\%{]}\end{tabular}} \\ \midrule
No recurrence no SFA & 0 & 0 & 1.1M / 1.7M & 30.7 & 28.7 \\
Recurrence only & 0.5 & 0 & 1.5M / 2.1M & 23.6 & 20.6 \\
SFA only & 0 & 0.5 & 1.1M / 1.7M & 25.1 & 22.9 \\
Recurrence and SFA & 0.5 & 0.5 & 1.5M / 2.1M & \textbf{21.2} & \textbf{19.2} \\
\bottomrule
\end{tabular}
}
\end{table}

\subsubsection{Supplementary findings}

We observe in Supplementary Table S1 that decreasing the simulation time step does not affect the performance. Although making the simulation of spiking dynamics more realistic, one might have anticipated that backpropagating through more time steps could hinder the training and worsen the performance as observed in standard RNNs often suffering from vanishing or exploding gradients \citep{Bengio1994}. With inputs ranging from 1,000 to over 7,000 steps using 1 ms intervals on TIMIT, our results demonstrate a promising scalability of surrogate gradient SNNs for processing longer sequences. 

Secondly, as reported in Supplementary Table S2, we did not observe substantial improvement when increasing the number of auditory nerve fibers past $\sim$5,000, even though there are approximately 30,000 of them in the human auditory system. This could be due to both the absence of a proper model for cochlear and hair cell processing in our pipeline and to the relatively low number of neurons ($<$1,000) in the subsequent layer. 

As detailed in Supplementary Table S3, reduced feedforward connectivity in the SNN led to poorer overall performance. This contrasts with our earlier findings on recurrent connectivity, highlighting the distinct functional roles of feedforward and feedback mechanisms in the network. 

Additionally, we incorporated trainable delays by replacing the fully connected feedforward matrices with dilated convolutions over the temporal dimension. While including delays resulted in similar overall performance, using the groups parameter to control connectivity led to more parameter-efficient models, as shown in Supplementary Table S4. This method of reducing connectivity proved more effective than our previous approach of randomly masking fully connected matrices. 

Finally, Supplementary Table S5 shows our results when using the more popular moving threshold formulation of SFA instead of the AdLIF model. Consistent with our previous findings on speech command recognition \citep{Bittar2022a}, the AdLIF implementation of SFA outperforms its moving threshold alternative, with the same number of parameters.

\subsection{Oscillation analysis}

Based on our previous architectural results that achieved satisfactory speech recognition performance using a physiologically inspired model, we hypothesise that the spiking dynamics of a trained network should, to some extent, replicate those occurring throughout the auditory pathway. Our investigation aims to discern if synchronisation phenomena resembling brain rhythms manifest within the implemented SNNs as they process speech utterances to recognise phonemes. 

\subsubsection{Synchronised gamma activity produces low-frequency rhythms}

\begin{figure}[ht]
    \centering
    \includegraphics[width=1.0\columnwidth]{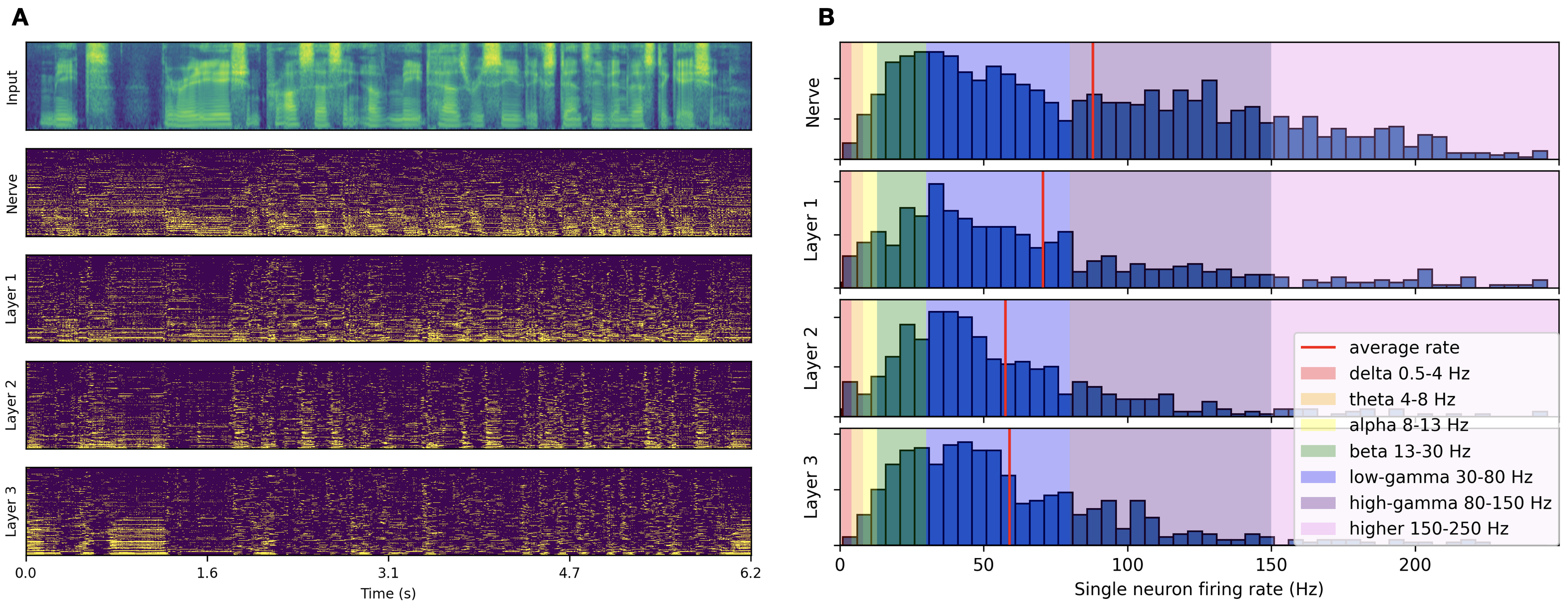}
    \caption{Spiking activity in response to speech input. \textbf{(A)} Input filterbank features and resulting spike trains produced across layers. For each layer, the neurons are vertically sorted on the y-axis by increasing average firing rate (top to bottom). The model uses a 2 ms time step, 16 CNN channels, 3 layers of size 512, 50\% AdLIF neurons, 100\% feedforward and 50\% recurrent connectivity with Dale's law. \textbf{(B)} Corresponding distribution of single neuron firing rates.}
    \label{fig:activity_speech}
\end{figure}

As illustrated in Fig. \ref{fig:activity_speech} (A), the spike trains produced by passing a test-set speech utterance through the trained architecture exhibit distinct low-frequency rhythmic features in all layers. By looking at the histogram of single neuron firing rates illustrated in Fig. \ref{fig:activity_speech} (B), we observe that the distribution predominantly peaks at gamma range, with little to no activity below beta. This reveals that the low-frequency oscillations visible in Fig. \ref{fig:activity_speech} (A) actually emerge from the synchronisation of gamma-active neurons. The resulting low-frequency rhythms appear to follow to some degree the intonation and syllabic contours of the input filterbank features and to persist across layers. Compared to the three subsequent layers, higher activity in the auditory nerve comes from the absence of inhibitory SFA and recurrence mechanisms. These initial observations suggest that the representation of higher-order features in the last layer is temporally modulated by lower level features already encoded in the auditory nerve fibers, even though each layer is seen to exhibit distinct rhythmic patterns. In the next section, we focus on measuring this modulation more rigorously via PAC analysis.

\subsubsection{Phase-amplitude couplings within and across layers}

\begin{figure}[ht]
    \centering
    \includegraphics[width=1.0\columnwidth]{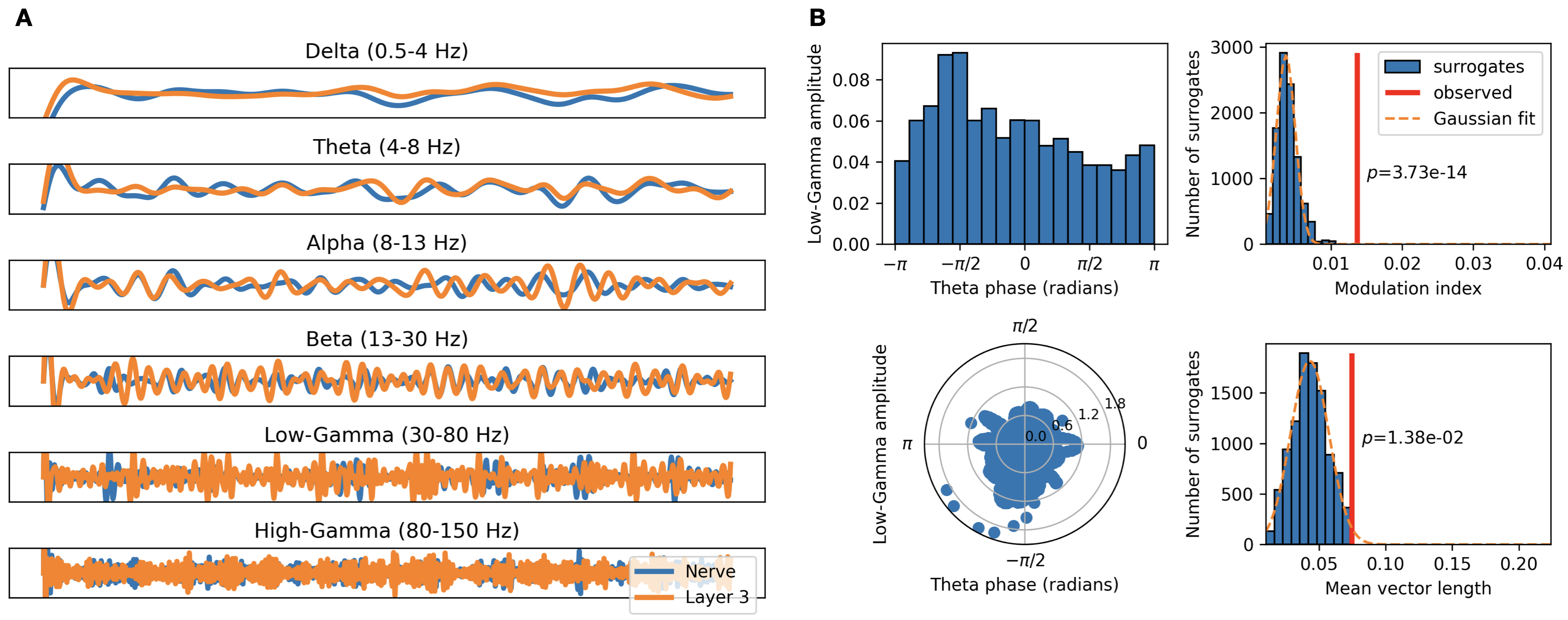}
    \caption{Cross-frequency coupling of population aggregated activity. \textbf{(A)} Population signals of auditory nerve fibers (blue) and last layer (orange) filtered in distinct frequency bands. \textbf{(B)} Modulation index and mean vector length metrics as measures of PAC between the theta band of the auditory nerve fibers and the low-gamma band of the last layer.}
    \label{fig:cfc}
\end{figure}

By aggregating over the relevant spike trains, we compute distinct EEG-like population signals for the auditory nerve fibers and each of the three SNN layers. These are then filtered in the different frequency bands, as illustrated in Fig. \ref{fig:cfc} (A), which allows us to perform CFC analyses. We measure PAC within-layer and across-layers between all possible combinations of frequency bands and find multiple significant forms of coupling for every utterance. 

An example of significant theta low-gamma coupling between the auditory nerve fibers and the last layer is illustrated in Fig. \ref{fig:cfc} (B). Here input low-frequency modulation is observed to significantly modulate the output gamma activity. This indicates that the network integrates and propagates intonation and syllabic contours across layers through synchronised neural activity along these perceptual cues. 

On the majority of utterances, we found significant CFCs between the input waveform and the population signal of the last layer. It is important to note that the synchronisation of neural signals to the auditory envelope emerged without imposing any theta or gamma activity in our network. The optimisation of the PER combined with sufficiently realistic spiking neuronal dynamics therefore represent sufficient conditions to reproduce some broad properties of neural oscillations observed in the brain, suggesting a general functional role of facilitating information processing. 

The activity of the final layer of the SNN stands out as the most significantly modulated overall. By architectural construction, modulation in that final layer has the greatest impact on the ASR task, as the spike trains from this layer are converted to phoneme probabilities using a small ANN. A higher number of couplings in the final layer correlates with a decrease in the PER. This suggests that CFCs may be associated with more selective integration of phonetic features, enhanced attentional processes, as well as improved assimilation of contextual information.

Alpha-band oscillations were the most frequently measured, consistent with biological evidence that the alpha rhythm is the most prominent oscillation in spontaneous EEG \citep{Berger1929}. 

More generally, the patterns of coupling between different neural populations and frequency bands were found to differ from one utterance to another. These variations indicate that the neural processing of our network is highly dynamic and depends on the acoustic properties and linguistic content of the input. The observed rich range of intra- and inter-layer couplings suggests that the propagation of low-level input features such as intonation and syllabic rhythm is only one aspect of these synchronised neural dynamics.

\subsubsection{Impact of Dale's law on oscillations}

As illustrated in Table \ref{table:oscillations}, SNNs that satisfy Dale's law display significantly higher numbers of CFCs. In biological neural networks, this principle contributes to a more structured and organised form of network dynamics, as each neuron is either excitatory or inhibitory but not both. When applying Dale's principle to SNNs, it constrains the network with more defined roles for neurons, leading to more coherent oscillatory patterns and more CFCs. 

In this study, we use equivalent models for both excitatory and inhibitory neurons, and simply allow them to adapt their parameters within the same fixed ranges during training. We observe that trained values of both membrane and adaptation time constants $\tau_u$ and $\tau_w$ converge to lower average values across the inhibitory populations compared to the excitatory ones. This suggests that, even with initially equivalent models, the network naturally differentiates the dynamics of excitatory and inhibitory neurons to fulfill their distinct functional roles. Future work could focus on better defining these two types of neurons from the outset, incorporating more biologically plausible initial parameter ranges.

When using Dale's law, the fixed excitatory-inhibitory (E/I) ratio plays a crucial role in shaping the neuronal dynamics inside the SNN. In our study, we observed that increasing the proportion of inhibitory neurons (E/I = 0.33) resulted in a similar ASR performance ($\sim 1\%$ absolute PER difference) compared to the standard ratio (E/I = 1), but with a reduced average firing rate of 56 Hz instead of 68 Hz for equivalent models with 50\% of neurons with SFA and 50\% of recurrent connectivity (see Table \ref{table:oscillations}). This finding suggests that a higher ratio of inhibitory neurons can achieve comparable task performance while maintaining a lower overall level of network activity.

\begin{table}[ht]
\caption{Effect of Dale's law, SFA, recurrence and delays on oscillatory activity. Comparison of the oscillatory activity resulting from passing the 64 longest TIMIT test-set utterances through different types of trained networks. The last two columns show the total number of significant PACs summed across all 64 utterances and frequency bands, for intra- and inter-layer relations respectively. All networks use a 2 ms time step, 16 CNN channels, 3 layers, 512 neurons per layer and 100\% feedforward connectivity (i.e., groups=1 for delays). The E/I ratio is 1 for models with Dale's law, except the last one where it is 0.33.}
\label{table:oscillations}
\vspace{0.5cm}
\centering
\resizebox{\textwidth}{!}{%
\begin{tabular}{llcccc}
\toprule
\multicolumn{1}{l}{\textbf{Model type}} &
  \textbf{Model configuration} &
  \textbf{\begin{tabular}[c]{@{}c@{}}Test PER \\ $[\boldsymbol{\%}]$\end{tabular}} &
  \textbf{\begin{tabular}[c]{@{}c@{}}Firing rate\\ $[$Hz$]$\end{tabular}} &
  \textbf{\begin{tabular}[c]{@{}c@{}}Number of intra- \\ layer PACs\end{tabular}} &
  \textbf{\begin{tabular}[c]{@{}c@{}}Number of inter- \\ layer PACs\end{tabular}} \\ \midrule
\multicolumn{1}{l}{AdLIF baseline}   & No recurrence no SFA & 26.9 & 98  & 133 & 192 \\
                                     & SFA only             & 23.7 & 72  & 213 & 390 \\
                                     & Recurrence only      & 21.0 & 76  & 180 & 132 \\
                                     & Recurrence and SFA   & 20.5 & 71  & 265 & 177 \\[+0.5ex] \hline\\[-1.5ex]
AdLIF with Dale                      & No recurrence no SFA & 30.7 & 100 & 513 & 643 \\
                                     & SFA only             & 25.1 & 77  & 564 & 603 \\
                                     & Recurrence only      & 23.6 & 67  & 432 & 278 \\
                                     & Recurrence and SFA   & 21.2 & 68  & 329 & 290 \\
                                     & Recurrence and SFA, E/I=0.33   & 22.3 & 56  & 526 & 496 \\[+0.5ex]  \hline]\\[-1.5ex]
AdLIF with delays                    & No recurrence no SFA & 26.2 & 92  & 116 & 54  \\
                                     & SFA only             & 23.3 & 77  & 91  & 53  \\
                                     & Recurrence only      & 21.6 & 68  & 205 & 66  \\
                                     & Recurrence and SFA   & 20.8 & 68  & 253 & 60  \\[+0.5ex] \hline\\[-1.5ex]
\multicolumn{1}{l}{Moving threshold} & SFA only             & 26.3 & 56  & 136 & 131 \\
\multicolumn{1}{l}{}                 & Recurrence and SFA   & 22.1 & 62  & 182 & 135 \\ \bottomrule
\end{tabular}}
\end{table}

\subsubsection{Impact of spike-frequency adaptation and recurrent connections on oscillations}

Across all model types (AdLIF or moving threshold SFA, with or without delays and Dale's law), both SFA and recurrent connections had an overall inhibitory effect, typically reducing the average network firing rate from around 100 Hz to roughly 60 Hz (see Table \ref{table:oscillations}). 

This regularisation of the network activity appears to enable more effective parsing and encoding of speech information, as it reliably led to improved ASR performance. While both forms of feedback exhibit an overall inhibitory effect, SFA operates at the individual neuron level whereas recurrent connections act at the layer level. 

SFA is known to encourage and stabilise the synchronisation of cortical networks \citep{Crook1998} and to promote periodic signal propagation \citep{Augustin2013}. In our results, this effect is pronounced in SNNs without Dale's law, where the inclusion of SFA was consistently associated with a significant increase in the number of measured CFCs. However, when Dale's law is applied, the overall number of CFCs is significantly higher with no noticeable impact of SFA on CFCs. This suggests that the stricter constraints imposed by Dale's law may lead to more uniform behaviour in CFCs, thereby reducing the observed influence of SFA.

In models with and without Dale's law, incorporating recurrent connections was consistently associated with a decrease in the number of inter-layer couplings, indicating more localised synchronisation. 

Finally, using the moving threshold formulation of SFA produces a lower firing rate and fewer significant PACs compared to our AdLIF baseline. The lower number of PACs might mean that the moving threshold formulation is less effective than the AdLIF at coordinating these interactions, which could explain the inferior task performance. Nevertheless, the discrepancy between the two approaches could potentially be narrowed with further hyperparameter optimisation.

\subsubsection{Impact of delays on oscillations}

As illustrated in Table \ref{table:oscillations}, SNNs with delays produced significantly fewer PACs, especially for inter-layer couplings, compared to the baseline with fully connected feedforward matrices. Introducing trainable delays through dilated convolutions allows for temporal dispersion of signals, which may desynchronise neural activity and explain the observed reduction in CFCs.

\subsubsection{Effects of training and input type on neuronal activity}
In order to further understand the emergence of coupled signals, we consider passing speech through an untrained network, as well as passing different types of noise inputs through a trained network. 

\begin{figure}[ht]
    \centering
    \includegraphics[width=1.0\columnwidth]{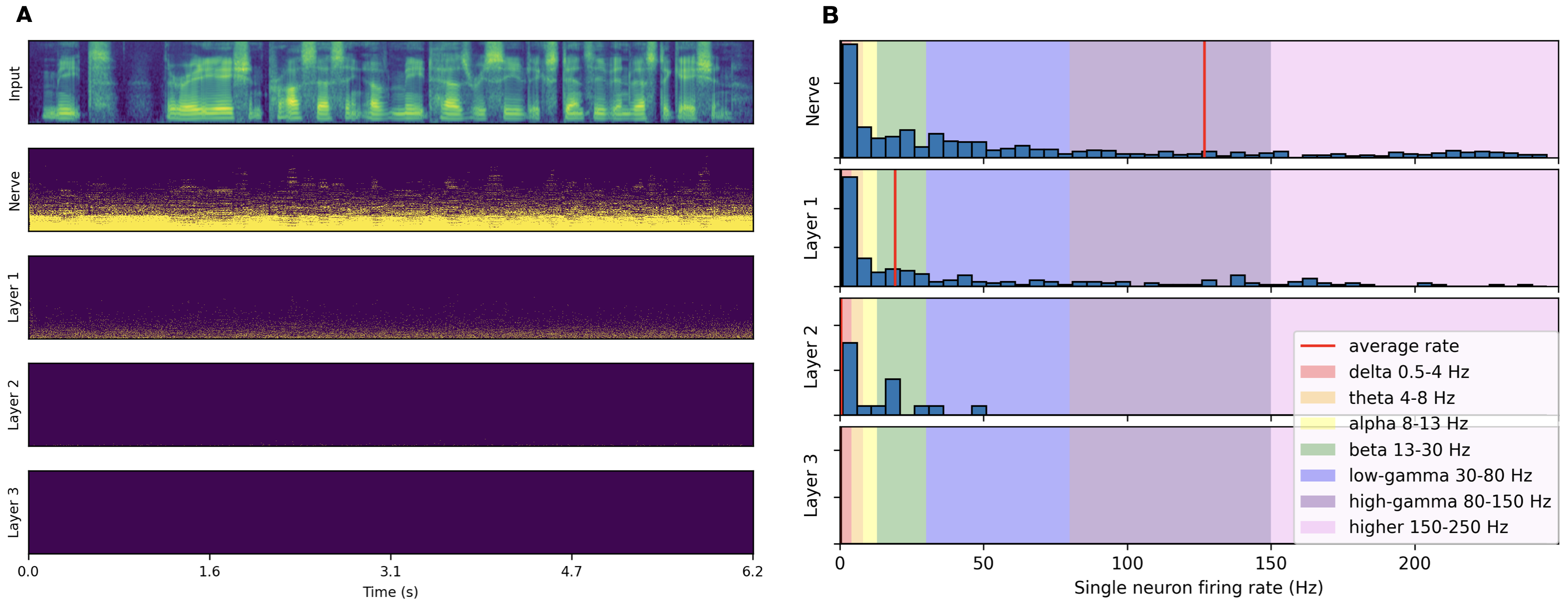}
    \caption{Spiking activity in an untrained network in response to speech input. \textbf{(A)} Input filterbank features and resulting spike trains produced across layers. The model uses a 2 ms time step, 16 CNN channels, 3 layers of size 512, 50\% AdLIF neurons, 100\% feedforward and 50\% recurrent connectivity. \textbf{(B)} Corresponding single neuron firing rate distributions.}
    \label{fig:activity_untrained}
\end{figure}

Trained architectures exhibit persisting neuronal activity across layers compared to untrained ones, where the activity almost completely decays after the first layer, as illustrated in Figure \ref{fig:activity_untrained}. This decay across layers persists even when increasing the input magnitude up to saturating auditory nerve fibers. This phenomenon can be attributed to the random weights in untrained architectures, which transform structured input patterns into uncorrelated noise, leading to vanishing neuronal activity in deeper layers. Our CFC analysis shows no significant coupling, even in layers with sufficient spiking activity, i.e., within the auditory nerve population and the first layer.

\begin{figure}[ht]
    \centering
    \includegraphics[width=1.0\columnwidth]{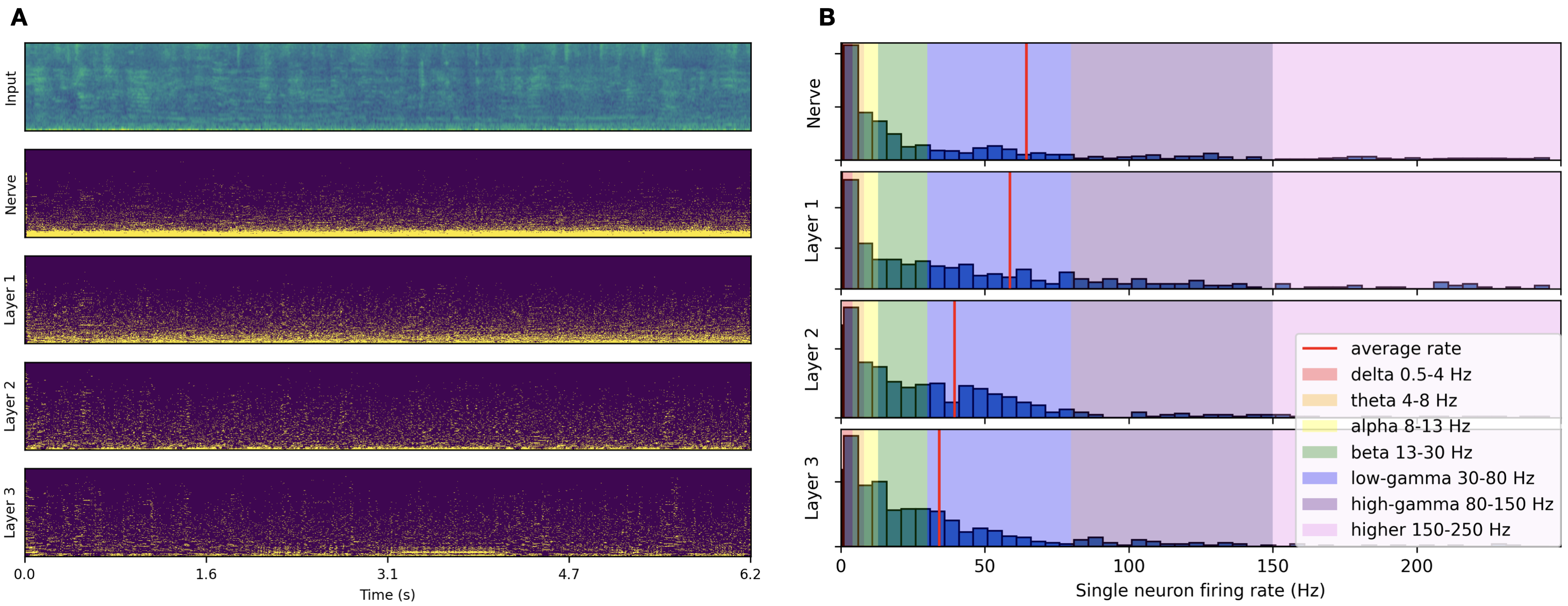}
    \caption{Spiking activity in response to babble noise input. \textbf{(A)} Input filterbank features and resulting spike trains produced across layers. The model uses a 2 ms time step, 16 CNN channels, 3 layers of size 512, 50\% AdLIF neurons, 100\% feedforward and 50\% recurrent connectivity. \textbf{(B)} Corresponding single neuron firing rate distributions.}
    \label{fig:activity_babble}
\end{figure}

In trained networks, noise inputs lead to single neuron firing rate distributions peaking at very low rates and where the activity gradually decreases across layers, as illustrated in Fig. \ref{fig:activity_babble} (B). This contrasts with the response to speech inputs seen in Fig. \ref{fig:activity_speech} (B) where the activity was sustained across layers with most of the distribution in the gamma range. We tested uniform noise as well as different noise sources (air conditioner, babble, copy machine and typing) from the the MS-SNSD dataset \citep{Reddy2019}. Compared to a speech input, all noise types yielded reduced average firing rates (from 60 Hz to about 40 Hz) with most of the neurons remaining silent. This highly dynamic processing of information is naturally efficient at attenuating its activity when processing noise or any input that does not induce sufficient synchronisation. Interestingly, babble noises were found in certain cases to induce some significant PAC patterns, whereas other noise types resulted in no coupling at all. Even though babble noises resemble speech and produced some form of synchronisation, they only triggered a few neurons per layer (see Fig. \ref{fig:activity_babble}). Overall, we showed that synchronicity of neural oscillations in the form of PAC results from training and is only triggered when passing an appropriate speech input. 

\subsubsection{Scaling to a larger dataset}

Our approach was extended to the Librispeech dataset \citep{Panayotov2015} with 960 hours of training data. After 8 epochs, we reached 9.5$\%$ word error rate on the \textit{test-clean} data split. As observed on TIMIT, trained models demonstrated similar CFCs in their spiking activity.

\subsubsection{Training on speech command recognition task}
With our experimental setup, the encoder is directly trained to recognise phonemes on TIMIT and subwords on Librispeech. One could therefore assume that the coupled gamma activity emerges from that constraint. In order to test this hypothesis, we run additional experiments on a speech command recognition task where no phoneme or subword recognition is imposed by the training. Instead the model is directly trained to recognise a set of short words. We use the same architecture as on TIMIT, except the average pooling layer is replaced by a readout layer as defined in \citet{Bittar2022a} which reduces the temporal dimension altogether, as required by the speech command recognition task. Interestingly, using speech command classes as ground truths still produces significant PAC patterns, especially in the last layer. These results indicate that the emergence of the studied rhythms does not require phoneme-based training and may be naturally emerging from speech processing. 

Using the second version of the Google Speech Commands data set \citep{Warden2018} with 35 classes, we achieve a test set accuracy of 97.05\%, which, to the best of our knowledge, improves upon the current state-of-the-art performance using SNNs of 95.35\% \citep{Hammouamri2023}. 

\section{CONCLUSION}

In this study, we introduced a physiologically inspired speech recognition architecture, centred around an SNN, and designed to be compatible with modern deep learning frameworks. As set out in the introduction, we first explored the capabilities and scalability of the proposed speech recognition architecture before analysing neural oscillations. 

Our preliminary architectural analysis demonstrated a satisfactory level of scalability to deeper and wider networks, as well as to longer sequences and larger datasets. This scalability was achieved through our approach of utilising the surrogate gradient method to incorporate an SNN into an end-to-end trainable speech recognition pipeline. In addition, our ablation experiments emphasised the importance of including SFA within the neuron model, along with layer-wise recurrent connections, to attain optimal recognition performance. Notably, our implementation of SFA using the AdLIF model outperformed the more popular moving threshold formulation, which corroborates our previous results on speech command recognition \citep{Bittar2022a}.

The subsequent analysis of the spiking activity across our trained networks in response to speech stimuli revealed that neural oscillations, commonly associated with various cognitive processes in the brain, did emerge from training an architecture to recognise words or phonemes. Through CFC analyses, we measured similar synchronisation phenomena to those observed throughout the human auditory pathway. During speech processing, trained networks exhibited several forms of PAC, including delta-gamma, theta-gamma, alpha-gamma, and beta-gamma, while no such coupling occurred when processing pure background noise. Our networks' ability to synchronise oscillatory activity in the last layer was also associated with improved speech recognition performance, which points to a functional role for neural oscillations in auditory processing. Even though we employ gradient descent training, which does not represent a biologically plausible learning algorithm, our approach was capable of replicating natural phenomena of macro-scale neural coordination. By leveraging the scalability offered by deep learning frameworks, our approach can therefore serve as a valuable tool for studying the emergence and role of brain rhythms.

Building upon the main outcome of replicating neural oscillations, our analysis on SFA and recurrent connections emphasised their key role in actively shaping neural responses and driving synchronisation via inhibition in support of efficient auditory information processing. Our results point towards further work on investigating more realistic feedback mechanisms including efferent pathways across layers. More accurate neuron populations could also be obtained using clustering algorithms.

Further analysis incorporated Dale's law which constrains neurons to be exclusively excitatory or inhibitory. This physiologically inspired principle proved to be a crucial consideration as it significantly increased the number of measured oscillations.

Aside from the fundamental aspect of developing the understanding of biological processes, our research on SNNs also holds significance for the fields of neuromorphic computing and energy efficient technology. Our exploration of the spiking mechanisms that drive dynamic and efficient information processing in the brain is particularly relevant for low-power audio and speech processing applications, such as on-device keyword spotting. In particular, the absence of synchronisation in our architecture when handling background noise results in fewer computations, making our approach well-suited for always-on models.

\section*{Author Contributions}

AB designed and implemented the architecture, performed training and analysis experiments, did most of the literature search, and wrote the bulk of the manuscript. PG secured funding, supervised the experimental work, and assisted with literature and writing. All authors contributed to the article and approved the submitted version.

\section*{Funding}

This project received funding under NAST: Neural Architectures for Speech Technology, Swiss National Science Foundation grant 185010 (\url{https://data.snf.ch/grants/grant/185010}).

\section*{Data Availability Statement}

All datasets used to conduct our study are publicly available. The TIMIT dataset, along with relevant access information, can be found on the Linguistic Data Consortium website at \url{https://catalog.ldc.upenn.edu/LDC93S1}. The Librispeech and Google Speech Commands datasets are directly available at \url{https://www.openslr.org/12}. and \url{https://www.tensorflow.org/datasets/catalog/speech_commands} respectively.

\section*{Code availability}

To facilitate the advancement of spiking neural networks, we have made our code available open source at \url{https://github.com/idiap/sparse}.

\bibliographystyle{Frontiers-Harvard} 
\bibliography{test}

\end{document}


\onecolumn
\firstpage{1}

\title[Supplementary Material]{{\helveticaitalic{Supplementary Material}}}
\maketitle

\section{Appendix}

\subsection{Forward-Euler first-order exponential integrator method}
\label{ssec:euler}

Consider the following differential equation,
\begin{equation}
    \tau\,\dot{y}=-y + \mathcal{N}(y) + \tau\,\gamma\,s(t) \, ,
\end{equation}
where $\mathcal{N}(y)$ is a nonlinear function and $s(t)$ is a spike train defined as
\begin{equation}
    s(t)=\sum_f\delta(t-t^f)\, .
\end{equation}
Multiplying both sides by $\frac{1}{\tau}\exp(\frac{t}{\tau})$ and then integrating over $[t_n, t_{n+1}]$, where $t_n=n\,\Delta t$, yields
\begin{equation}
    \int_{t_n}^{t_{n+1}}\Big(\dot{y}\exp{\frac{t}{\tau}}+y\,\frac{1}{\tau}\,\exp{\frac{t}{\tau}}\Big)dt=\frac{1}{\tau}\int_{t_n}^{t_{n+1}}\mathcal{N}(y)\exp{\frac{t}{\tau}}\,dt+\gamma\sum_f\int_{t_n}^{t_{n+1}}\exp\frac{t}{\tau}\,\delta({t-t^f})\, .
\end{equation}
The left hand side has an exact solution
\begin{equation}
    y(t)\,\exp\frac{t}{\tau}\bigg|_{t=t_n}^{t_{n+1}}=\Big(y_{n+1}\exp\frac{\Delta t}{\tau}-y_n\Big)\exp\frac{t_n}{\tau}\, .
\end{equation}
For the first term of the right hand side, the nonlinearity $\mathcal{N}(y)$ can be approximated as constant over $[t_n, t_{n+1}]$ for sufficiently small $\Delta t$, so that $\mathcal{N}(y)\approx\mathcal{N}(y_n)$ and we can solve it as
\begin{equation}
    \frac{1}{\tau}\int_{t_n}^{t_{n+1}}\mathcal{N}(y)\exp{\frac{t}{\tau}}\,dt\approx\mathcal{N}(y_n)\,\exp\frac{t}{\tau}\bigg|_{t=t_n}^{t_{n+1}}=\mathcal{N}(y_n)\exp\frac{t_n}{\tau}\Big(\exp\frac{\Delta t}{\tau}-1\Big)\, .
\end{equation}
Finally for the last term, the width $\Delta t$ of the interval $[t_n, t_{n+1}]$ can be set sufficiently small to include at most a single spike. The exact firing time $t^f\in[t_n,t_{n+1}]$ can then be discretised as $t^f=t_n$ so that $s_n=\sum_f\delta(t_n-t^f)$ and
\begin{equation}
    \gamma\sum_f\exp\frac{t^f}{\tau}\bigg|_{t^f\in[t_n,t_{n+1}]}=\gamma\exp\frac{t^n}{\tau}\,s_n
\end{equation}
Putting everything together, we get the following update equation for $y$ in discrete time,
\begin{equation}
    y_{n+1}=\exp\frac{-\Delta t}{\tau}\Big(y_n+\gamma\,s_n\Big)+\Big(1-\exp\frac{-\Delta t}{\tau}\Big)\,\mathcal{N}(y_n) \, .
\end{equation}

\subsection{Eigenvalues of AdLIF free equations}
\label{ssec:eigenvalues}

The free equations of the AdLIF neuron model are obtained by considering Eqs. (3) and (4) in the special case where there is no input, $I(t)=0$, and no emitted spikes, $s(t)=0$. They can be rewritten in matrix form as,
\begin{equation}
    \frac{d}{dt} \begin{bmatrix}
        u \\
        w
    \end{bmatrix} = \begin{bmatrix}
        -1/\tau_u & -1/\tau_u \\
        a/\tau_w & -1/\tau_w 
    \end{bmatrix} \begin{bmatrix}
        u \\
        w
    \end{bmatrix}=A\,\begin{bmatrix}
        u \\
        w
    \end{bmatrix} \, .
\end{equation}
The eigenvalues can be found by setting the determinant of $A-\lambda\mathbb{I}$ to zero, 
\begin{equation}
    \begin{vmatrix}
        -1/\tau_u-\lambda & -1/\tau_u \\
        a/\tau_w & -1/\tau_w -\lambda\,\,
    \end{vmatrix} = 0 \, ,
\end{equation}
yielding the characteristic polynomial,
\begin{equation}
    \lambda^2+\lambda\Bigg(\frac{1}{\tau_u}+\frac{1}{\tau_w}\Bigg)+\frac{1+a}{\tau_u\tau_w}=0\, ,
\end{equation}
whose roots correspond to the two eigenvalues of the system,
\begin{equation}
\label{eq_eigs}
    \lambda_{1,2}=-\frac{1}{2}\Bigg(\frac{1}{\tau_u}+\frac{1}{\tau_w}\Bigg)\pm \frac{1}{2}\sqrt{\Bigg(\frac{1}{\tau_u}+\frac{1}{\tau_w}\Bigg)^2-\frac{4(1+a)}{\tau_u\tau_w}} \, .
\end{equation}
In order to prevent the occurrence of exponentially growing solutions and ensure stability, both eigenvalues need to have a strictly negative real part, which can be realised by imposing a lower bound $a>-1$ on the coupling strength. Moreover, allowing eigenvalues to have a nonzero imaginary part introduces the potential for oscillatory modes that may amplify perturbations. This could cause some challenges in terms of numerical stability, convergence and interpretability, especially in the context of deep neural networks trained with gradient descent. We therefore impose an additional upper bound on the values of $a$ leading to the overall stability condition,
\begin{equation}
    -1<a\leq\frac{\big(\tau_w-\tau_u\big)^2}{4\tau_u\tau_w} \, .
\end{equation}

\subsection{Kernel formulation of a spiking neuron}
\label{ssec:kernel}

Using the SRM formulation, the membrane potential $u(t)$ is described as,
\begin{equation}
    u(t)=\int_0^\infty\kappa(s)\,I(t-s)ds + \int_0^\infty\eta(s)\,S(t-s)ds \, ,
\end{equation}
where the two kernels $\kappa(s)$ and $\eta(s)$ describe the response to an input pulse and the response to an afterspike reset pulse respectively. It can be shown that the differential equations of a linear AdLIF neuron has an equivalent kernel formulation with,
\begin{subequations}
\begin{align}
\kappa(s)&=\Big(\beta_1\,e^{\lambda_1\,s}+\beta_2\,e^{\lambda_2\,s}\Big)\Theta(s) \\
\eta(s)&=\Big(\gamma_1\,e^{\lambda_1\,s}+\gamma_2\,e^{\lambda_2\,s}\Big)\Theta(s) \, ,
\end{align}
\end{subequations}
where $\lambda_1$, $\lambda_2$ are the eigenvalues of the system given in Supplementary Eq. \eqref{eq_eigs} and $\Theta(s)$ is the Heaviside step function. The coefficients $\beta_1$, $\beta_2$ of the input kernel are such that the membrane potential increases by $\Delta u=1$, without any effect on the recovery current, i.e., $\Delta w=0$,
\begin{equation}
    \beta_1=\frac{\tau_u\lambda_2+1}{\tau_u(\lambda_2-\lambda_1)} \quad\text{and}\quad \beta_2=1-\beta_1 \, .
\end{equation}
The coefficients $\gamma_1$, $\gamma_2$ on the afterspike reset kernel are such that the membrane potential decreases by $\Delta u=\vartheta-u_r$ and the recovery current jumps by an amount $\Delta w=b$,
\begin{equation}
    \gamma_1=\frac{b-(\vartheta-u_r)(\tau_u\lambda_2+1)}{\tau_u(\lambda_2-\lambda_1)} \quad\text{and}\quad \gamma_2=-(\vartheta-u_r)-\gamma_1 \, .
\end{equation}

\section{Supplementary Tables}

\begin{table}[ht]
\caption{Hyperparameter tuning for the simulation time step on the TIMIT dataset. PERs are reported after 50 epochs using 16 CNN channels, 3 layers of 512 neurons each, 50\% of AdLIF neurons, 100\% feedforward and 50\% recurrent connectivity.}
\label{table:timit_dt}
\vspace{0.5cm}
\centering
\begin{tabular}{lccc}
\toprule
\textbf{Time step [ms]} &
\textbf{Epoch duration [min]} &
\textbf{Test PER [\%]} &
\textbf{Validation PER [\%]} \\ \midrule
5 & 21 & 20.5 & 18.2 \\
2 & 53 & 20.4 & 18.7 \\
1 & 156 & 20.6 & 18.2 \\
\bottomrule
\end{tabular}
\end{table}

\begin{table}[h]
\caption{Hyperparameter tuning for the number of CNN channels on the TIMIT dataset. PERs are reported after 50 epochs using a 5 ms time step, 3 layers, 512 neurons per layer, 50\% of AdLIF neurons, 100\% feedforward and 50\% recurrent connectivity. Bold values indicate the lowest achieved PERs.}
\label{table:timit_cnn}
\vspace{0.5cm}
\centering
\resizebox{\textwidth}{!}{
\begin{tabular}{lcccc}
\toprule
\textbf{CNN channels} &
\textbf{Nerve fibers} &
\textbf{Parameters in complete encoder} &
\textbf{Test PER [\%]} &
\textbf{Validation PER [\%]} \\ \midrule
8 & 592 & 1.8M & 20.9 & 18.9 \\
16 & 1,184 & 2.1M & 20.5 & 18.2 \\
32 & 2,368 & 2.7M & 20.2 & 18.4 \\
64 & 4,736 & 3.9M & \textbf{19.8} & \textbf{18.0} \\
128 & 9,472 & 6.4M & 21.3 & 18.9 \\
\bottomrule
\end{tabular}
}
\end{table}

\begin{table}[h]
\caption{Hyperparameter tuning for the feedforward connectivity on the TIMIT dataset. PERs are reported after 50 epochs using a 5 ms time step, 16 CNN channels, 3 layers of 512 neurons each, 50\% of AdLIF neurons and 50\% recurrent connectivity. Bold values indicate the lowest achieved PERs.}
\label{table:timit_feedforward}
\vspace{0.5cm}
\centering
\begin{tabular}{lccc}
\toprule
\textbf{Feedforward connectivity} &
\textbf{Number of parameters} &
\textbf{Test PER [\%]} &
\textbf{Validation PER [\%]} \\ \midrule
0.2 & 629k / 1.2M & 22.0 & 19.6 \\
0.5 & 968k / 1.5M & 21.6 & 19.7 \\
0.8 & 1.3M / 1.8M & 20.7 & 18.8 \\
1.0 & 1.5M / 2.1M & \textbf{20.5} & \textbf{18.2} \\
\bottomrule
\end{tabular}
\end{table}

\begin{table}[h]
\caption{Hyperparameter tuning for trainable delays on the TIMIT dataset. PERs are reported after 50 epochs using a 5 ms time step, 16 CNN channels, 3 layers of 512 neurons. Bold values indicate the lowest achieved PERs.}
\label{table:timit_delays}
\vspace{0.5cm}
\centering
\resizebox{\textwidth}{!}{
\begin{tabular}{lcccc}
\toprule
\textbf{Model type} &
\textbf{Conv Groups} &
\textbf{Number of parameters} &
\textbf{Test PER [\%]} &
\textbf{Validation PER [\%]} \\ \midrule
No recurrence no SFA & 1 &  2.3M / 2.8M & 26.2 & 24.6 \\
Recurrence only & 1 & 2.7M / 3.2M & 21.6 & 19.5 \\
SFA only & 1 & 2.3M / 2.8M & 23.3 & 20.7 \\
Recurrence and SFA & 1 & 2.7M / 3.2M & \textbf{20.8} & 18.6 \\
Recurrence and SFA & 4 & 968k / 1.5M & 21.0 & \textbf{18.4} \\
\bottomrule
\end{tabular}
}
\end{table}

\begin{table}[h]
\caption{Results with moving threshold SFA on the TIMIT dataset. PERs are reported after 50 epochs using a 2 ms time step, 16 CNN channels, 3 layers of 512 neurons.}
\label{table:timit_movthres}
\vspace{0.5cm}
\centering
\begin{tabular}{lccc}
\toprule
\textbf{Model type} &
\textbf{Number of parameters} &
\textbf{Test PER [\%]} &
\textbf{Validation PER [\%]} \\ \midrule
SFA only & 1.1 / 1.7M & 26.3 & 23.9 \\
Recurrence and SFA & 1.5 / 2.1M & 22.1 & 19.2 \\
\bottomrule
\end{tabular}
\end{table}

\begin{table}[h]
\caption{Sensitivity to frequency of auditory nerve fibers. The second column gives the sensitivity ranges measured in \citet{Devi2022} that we compute from their reported mean Q10 values in the normal hearing group. The third column gives the Mel scale centers using 80 filters that surround the corresponding sensitivity ranges.}
\label{table:q10}
\vspace{0.5cm}
\centering
\resizebox{\textwidth}{!}{
\begin{tabular}{cccc}
\toprule
\textbf{\begin{tabular}[c]{@{}c@{}}Characteristic \\ {Frequency [Hz]}\end{tabular}} &
\textbf{\begin{tabular}[c]{@{}c@{}}Sensitivity \\ {range [Hz]}\end{tabular}} &
\textbf{\begin{tabular}[c]{@{}c@{}}Nearby Mel bin \\ {centers [Hz]}\end{tabular}} &
\textbf{\begin{tabular}[c]{@{}c@{}}Overlapping Number \\ {of bins}\end{tabular}}\\ \midrule
500 & 416-583 & 416, 452, 488, 525, 564, 604 & 5-6 \\
1000 & 881-1119 & 872, 921, 973, 1026, 1080, 1136 & 5-6 \\
2000 & 1770-2230 & 1729, 1806, 1886, 1967, 2052, 2139, 2228 & 6-7 \\
4000 & 3566-4434 & 3553, 3688, 3827, 3970, 4117, 4270, 4427 & 7 \\
6000 & 5501-6499 & 5479, 5674, 5875, 6083, 6297, 6519 & 5-6 \\
\bottomrule
\end{tabular}
}
\end{table}

\bibliographystyle{Frontiers-Harvard} 
\bibliography{test}